\documentclass[10pt,twocolumn,letterpaper]{article}

\usepackage{cvpr}
\usepackage{times}
\usepackage{epsfig}
\usepackage{graphicx}
\usepackage{amsmath}
\usepackage{amssymb}
\usepackage{subfigure}
\usepackage{multirow}
\usepackage{stfloats}
\usepackage{tabularx}
\usepackage{pifont}
\usepackage{makecell}
\usepackage{bbding}
\usepackage{threeparttable}
\usepackage{mathtools}
\usepackage[linesnumbered]{algorithm2e}

\usepackage[pagebackref=true,breaklinks=true,letterpaper=true,colorlinks,bookmarks=false]{hyperref}

\cvprfinalcopy 

\def\cvprPaperID{2657} 

\ifcvprfinal\fi
\begin{document}

\title{Adaptive NMS: Refining Pedestrian Detection in a Crowd}

\author{Songtao Liu$^{1,2,3}$ \quad Di Huang$^{1,2,3}$\thanks{corresponding author} \quad Yunhong Wang$^{1,3}$\\
$^1$Beijing Advanced Innovation Center for Big Data and Brain Computing, Beihang University\\
$^2$State Key Laboratory of Software Development Environment, Beihang University\\
$^3$School of Computer Science and Engineering, Beihang University, Beijing 100191, China\\
{\tt\small \{liusongtao, dhuang, yhwang\}@buaa.edu.cn}
}

\maketitle

\begin{abstract}
  Pedestrian detection in a crowd is a very challenging issue. This paper addresses this problem by a novel Non-Maximum Suppression (NMS) algorithm to better refine the bounding boxes given by detectors. The contributions are threefold: (1) we propose adaptive-NMS, which applies a dynamic suppression threshold to an instance, according to the target density; (2) we design an efficient subnetwork to learn density scores, which can be conveniently embedded into both the single-stage and two-stage detectors; and (3) we achieve state of the art results on the CityPersons and CrowdHuman benchmarks.
\end{abstract}

\section{Introduction}

During the last two decades, pedestrian detection, as a special branch of general object detection, has received considerable attention. In the literature, many solutions have been presented to handle such an issue, and similar as in general object detection, the past several years have witnessed its technical development from models relying on hand-crafted features \cite{ACF,ICF,LDCF,checkboard} to deep learning networks \cite{fasterBF,how-far,occluded-attention,repulsion,or_cnn}. Due to the capability of learning discriminative features, Convolutional Neural Networks (CNN) based approaches dominate this area, and the results on public benchmarks are significantly promoted.

\begin{figure}[t]
\begin{center}
   \includegraphics[width=0.95\linewidth]{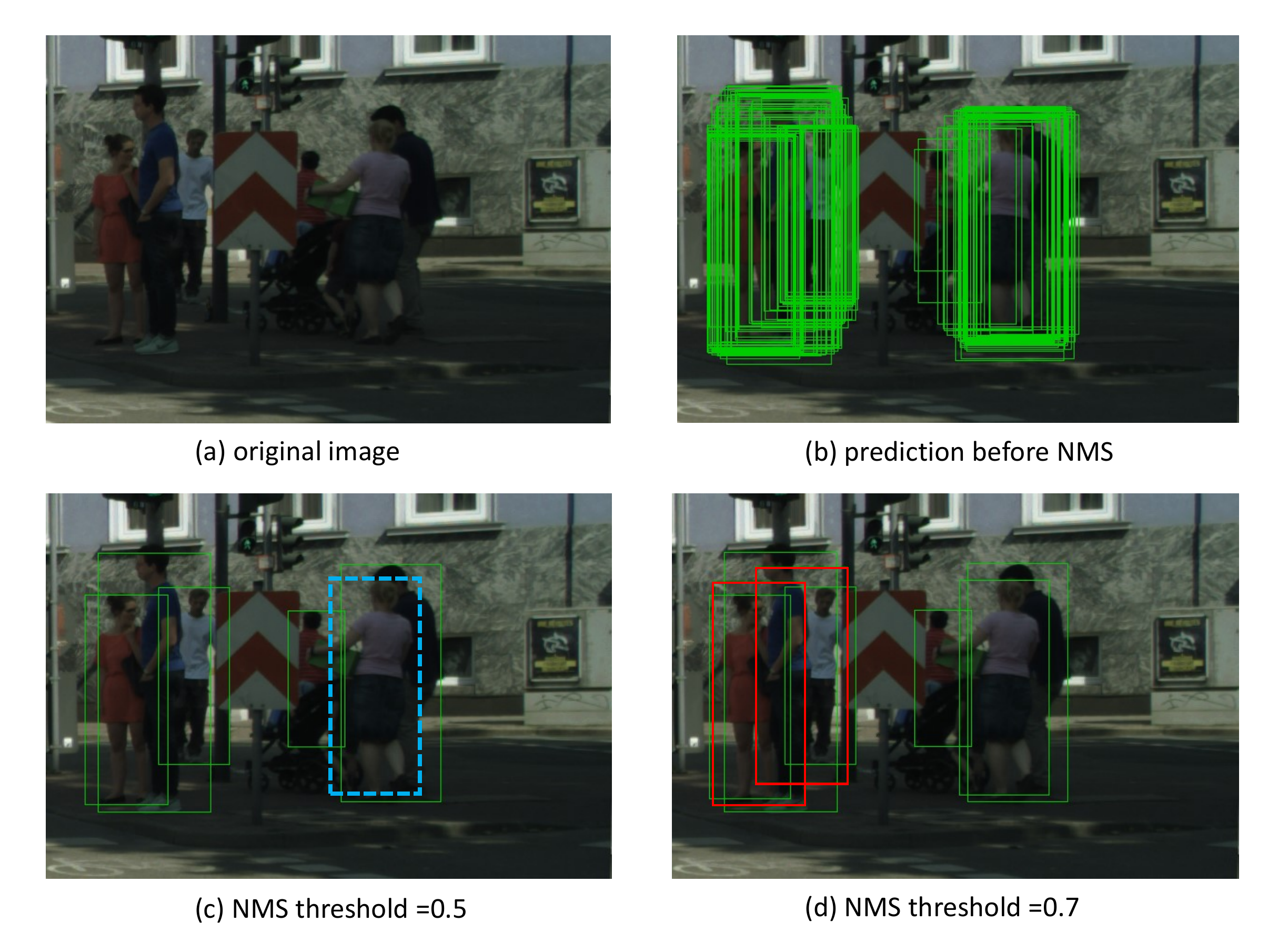}
\end{center}
   \caption{Illustration of greedy-NMS results of different thresholds. The blue box shows the missing object, while the red ones highlight false positives. The bounding boxes in (b) are generated using Faster R-CNN. In a crowd scene, a lower NMS threshold may remove true positives (c) while a higher NMS threshold may increase false positives (d). The threshold for visualization is above 0.3.}
\label{fig:NMS}
\vspace{-0.4cm}
\end{figure}

In recent years, pedestrian detection is urgently required in the real-world scenario where the density of people is high, \emph{i.e.}, airports, train stations, shopping malls \emph{etc}. Despite great progress achieved, detecting pedestrians in those scenes still remains difficult, evidenced by significant performance drops of state of the art methods.  For example, OR-CNN \cite{or_cnn}, a more recent work, reports a Miss Rate (MR) of 4.1\% on the Caltech database \cite{caltech}, which does not consider this challenge. Its MR degrades to 11.0\% on CityPersons \cite{citypersons}, where 26.4\% pedestrians are overlapped with an Intersection over Union (IoU) above 0.3 and the average of pairwise overlap between two human instances (larger than 0.5 IoU) is 0.32 per image. Therefore, it becomes a necessity to work on pedestrian detection in a crowd. While one may argue that this problem is the same as occlusion, they are indeed different, as in a crowd scene, pedestrians whose appearances are similar often occlude each other by a large part, making it even more challenging.

This work focuses on this issue, and we start with the analysis of deep learning based detectors. As we know, existing detectors either directly regress the default anchors into detection boxes on the feature maps (single-stage detectors, \emph{e.g.}, SSD~\cite{ssd}, YOLO~\cite{yolo,yolo9000}, RetinaNet~\cite{focal-loss}), or first generate category independent region proposals and then refine them (two-stage detectors, \emph{e.g.}, Faster R-CNN~\cite{FasterCNN}, R-FCN~\cite{R-FCN}). All the methods produce large numbers of false positives near the ground truth, and the greedy Non-Maximum Suppression (NMS) is necessary to screen out final detections by sharply reducing the false positives. In a crowded scenario, however, greedy-NMS encounters a problem. As shown in Fig.~\ref{fig:NMS}, even with a powerful detector that can predict exactly the same bounding boxes as the ground truth, the highly overlapped ones are still suppressed by the post process of greedy-NMS with a normal threshold of 0.5. It makes the current CNN based detectors confront with a dilemma for the single threshold of greedy-NMS: a lower threshold leads to missing highly overlapped objects while a higher one brings in more false positives.

To address this problem, \cite{repulsion} and \cite{or_cnn} propose additional penalties to produce more compact bounding boxes and thus become less sensitive to the threshold of NMS. The ideal solution for crowds under their pipelines with greedy-NMS is to set a high threshold to preserve highly overlapped objects and predict very compact (higher than the threshold) detection boxes for all instances to reduce false positives. Unfortunately, this is not so easy, as the CNN based detectors often assign correlated scores to the neighboring regions around the object.

Recently, \cite{soft_nms} proposes a soft version of NMS, which decreases the associated detection scores according to an increasing function of overlap instead of discarding them. There also exist some works~\cite{relation,learningNMS} that build an extra module or network to learn the NMS function from data. They show a better performance than greedy-NMS in general object detection. In contrast, in a crowded scenario, the NMS function has to process a much larger set of highly-overlapped boxes and a considerable part of them are true positives. While similar softer heuristics or learning methods may also be applied, they are inefficient as soft-NMS still blindly penalizes highly overlapped boxes. Furthermore, the similarity of CNN based appearance features blurs the boundaries between highly overlapped true positives and duplicates. \cite{QUBO} presents a quadratic unconstrained binary optimization solution to replace the greedy NMS in pedestrian detection, but it also sets a hard threshold to suppress all highly-overlapped detection boxes like greedy-NMS. \cite{ddp} extends the optimization model with individualness scores, which relies on discriminative CNN features.

In this paper, we propose a new NMS algorithm named adaptive-NMS that acts as a more effective alternative to deal with pedestrian detection in a crowd. Intuitively, a high NMS threshold keeps more crowded instances while a low NMS threshold wipes out more false positives. The adaptive-NMS thus applies a dynamic suppression strategy, where the threshold rises as instances gather and occlude each other and decays when instances appear separately. To this end, we design an auxiliary and learnable sub-network to predict the adaptive NMS threshold for each instance.

Experiments are conducted on the CityPersons \cite{citypersons} and CrowdHuman~\cite{crowdhuman} databases, and our adaptive-NMS delivers promising improvements for both the two-stage and single-stage detectors on crowded pedestrian detection, indicating its effectiveness. Additionally, we reach state of the art performance, \emph{i.e.} 10.8\% MR$^{-2}$ on CityPersons and 49.73\% MR$^{-2}$ on CrowdHuman.

\section{Related Work}

\textbf{Generic object detection.} The traditional approaches to object detection are based on sliding window or region proposal classification using hand-crafted features. In the era of deep learning, R-CNN~\cite{RCNN}, builds the two-stage framework by combining the straightforward strategy of box proposal generation like SS~\cite{selective-search} and a CNN based classifier on these region candidates and displays a breathtaking improvement. Its descendants (\emph{e.g.}, Fast R-CNN \cite{fastCNN}, Faster R-CNN \cite{FasterCNN}) update the two-stage framework and achieve dominant performance. In contrast to the two-stage approaches, another alternative is single-stage framework based (\emph{e.g.}, SSD~\cite{ssd}, YOLO~\cite{yolo,yolo9000}), which skips the proposal generation step and directly predicts bounding boxes and class probabilities on deep CNN features, aiming to accelerate detection.

\textbf{Pedestrian detection.} Traditional pedestrian detectors, such as ACF \cite{ACF}, LDCF \cite{LDCF} and Checkerboards \cite{checkboard}, extend the Viola and Jones paradigm \cite{VJ} to exploit various filters on Integral Channel Features (ICF) \cite{ICF} with the sliding window strategy.

Afterward, coupled with the prevalence of deep learning techniques, CNN-based models rapidly dominate this field. In \cite{fasterBF}, hand-crafted features are replaced with deep neural network features before being fed into a boosted decision forest. \cite{MSCNN} performs detection at multiple layers to match objects of different scales, and adopts an upsampling operation to handle small instances.
\cite{what-can} presents a jointly learning framework with extra features to further improve performance.
\cite{ALF} explores the potential of single-stage detectors on pedestrian detection by stacking multi-step prediction for asymptotic localization.

For the occlusion issue, many efforts have been made in the past years. A common framework~\cite{part1,part2,part3,part4,part5,bi-box} for occlusion handling is to learn a series of part detectors and integrate the results to localize occluded pedestrians. More recently, several works \cite{how-far,occluded-attention,haikang,repulsion,or_cnn} focus on a more challenging issue of detecting pedestrian in a crowd. \cite{citypersons} and \cite{crowdhuman} propose two pedestrian datasets (\emph{i.e.}, CityPersons and CrowdHuman) to better evaluate detectors in crowd scenarios. \cite{occluded-attention} employs an attention mechanism across channels to represent various occlusion patterns. \cite{haikang} operates  somatic topological line localization to reduce ambiguity. \cite{repulsion} introduces a bounding box regression loss to not only push each proposal to reach its designated target, but also keep it away from other surrounding objects. Similarly, \cite{or_cnn} designs an aggregation penalty to enforce the proposals locate closely and compactly to the ground-truth objects. These two works \cite{repulsion,or_cnn} ameliorate detectors to produce more compact proposals and thus become less sensitive to the threshold of NMS in crowded scenes. Another interesting attempt \cite{end-to-end} uses a recurrent LSTM to sequentially generate detections without NMS, but this detection pipeline suffers from scale variations.

\textbf{Non-Maximum Suppression.} NMS is a widely used post process algorithm in computer vision. It is an essential component of many detection methods, such as edge detection \cite{edge-nms}, feature point detection \cite{point-nms} and object detection \cite{FasterCNN,FPN,focal-loss}. Moreover, despite significant progress in general object detection by deep learning, the hand-crafted and greedy NMS is still the most effective method for this task.

Recently, soft-NMS \cite{soft_nms} and learning NMS \cite{learningNMS} are proposed to improve NMS results. Instead of discarding all the surrounding proposals with the scores below the threshold, soft-NMS lowers the detection scores of neighbors by an increasing function of their overlap with the higher scored bounding box. It is conceptually satisfying, but still treats all highly-overlapped boxes as false positives. \cite{learningNMS} attempts to learn a deep neural network to perform the NMS function using only boxes and their scores as input, but the network is specifically designed and very complex. \cite{relation} proposes an object relation module to learn the NMS function as an end-to-end general object detector.  \cite{fitnessNMS} and \cite{iounet} replace the classification scores of proposals used in the NMS process with learned localization confidences to guide NMS to preserve more accurately localized bounding boxes. These methods prove effective in general object detection, but as we state, pedestrian detection in a crowd has its own challenge. Therefore, different from them, we propose to learn the density around each ground truth object as its own suppression threshold, sharing some similarity with the crowd density map estimation in the people counting task \cite{count1,count2}. It reduces the requirement for instance-discriminative CNN features, which is the major issue in the crowd scene.

To address pedestrian detection in a crowd, \cite{QUBO} proposes a quadratic unconstrained binary optimization solution to suppress detection boxes, which uses detection scores as a unary potential and overlaps between detections as a pairwise potential to produce final results. But it still applies a hard threshold to blindly suppress detection boxes as greedy-NMS does. \cite{ddp} adopts the determinantal point process based optimal model with additional individualness scores to discriminate different pedestrians. However, as detectors pay less attention to intra-class differences, the CNN features for crowded individuals tend to be less discriminative, and its optimization procedure also consumes more time. As a result, how to robustly process detection proposals in crowded scenarios is still one of the most critical issues for pedestrian detection.

\section{Method}
\begin{figure}[t]
\begin{center}
   \includegraphics[width=0.90\linewidth]{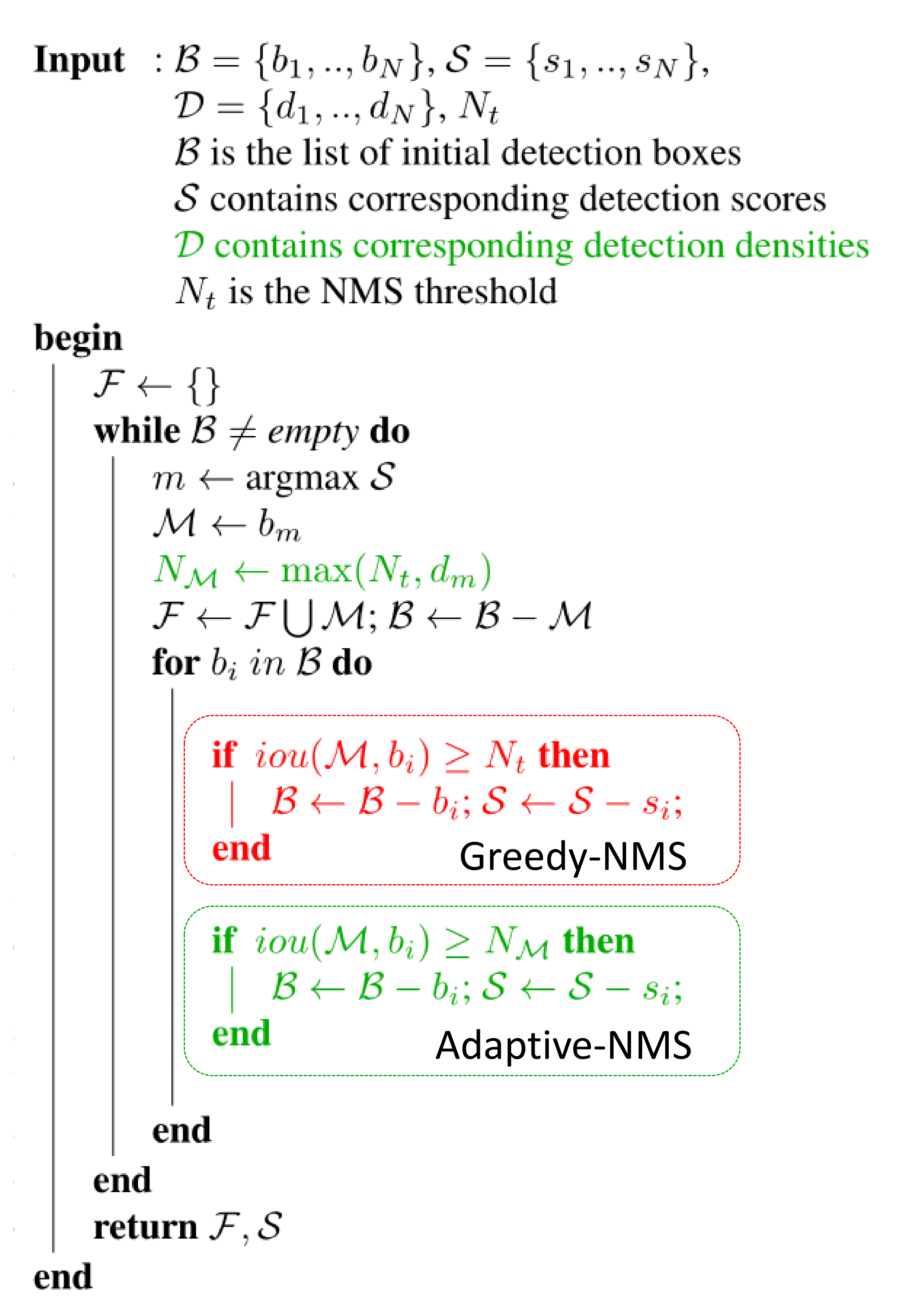}
\end{center}
\vspace{-0.35cm}
   \caption{The pseudo code in red is replaced by that in green in adaptive-NMS, which adaptively suppresses the detections by scaling their NMS threshold according to their densities.}
\label{fig:algorithm}
\vspace{-0.35cm}
\end{figure}

\begin{figure*}[t]
\begin{center}
   \includegraphics[width=0.95\linewidth]{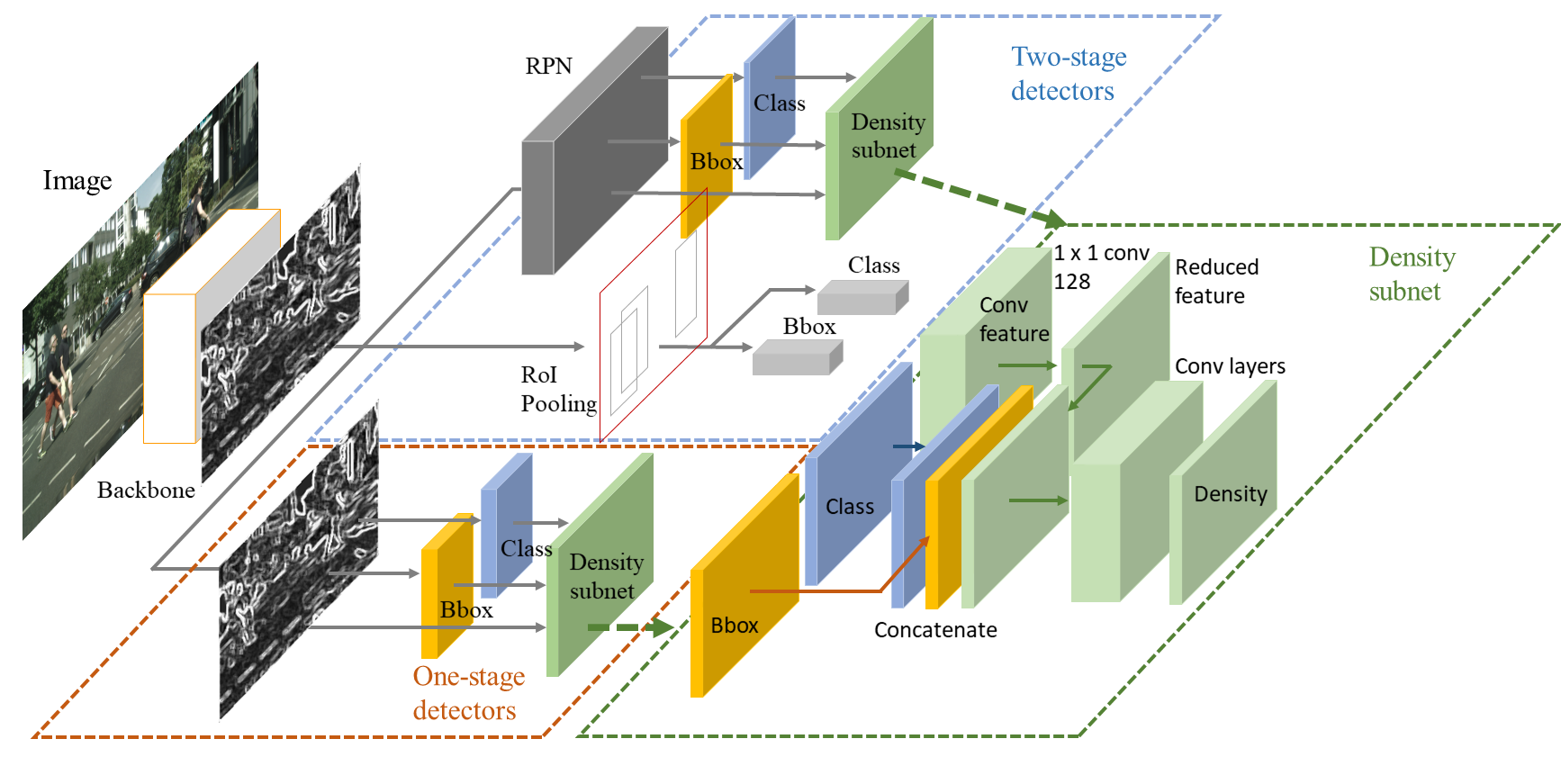}
\end{center}
\vspace{-0.4cm}
   \caption{Density prediction framework for both the two-stage and one-stage detectors. We add the density prediction subnet on the top of RPN for two-stage detectors, taking the objectness predictions, bounding box predictions and conv features as input. For one-stage detectors, the subnet is deployed behind the final detection network in a similar way.}
\label{fig:detector}
\vspace{-0.2cm}
\end{figure*}
\subsection{Greedy-NMS Revisit}
In pedestrian detection, the commonly used detection evaluation metric is log-average Miss Rate on False Positive Per Image (FPPI) in $[10^{-2}, 10^0]$ (denoted as MR or MR$^{-2}$ following \cite{caltech}), where the overlap criterion for a true positive is usually 0.5. MR is a good indicator for the detectors applied in the real-world applications since it shows the ability of the detector for balancing recall and precision. As shown in Fig.~\ref{fig:algorithm}, starting with a set of detection boxes $\mathcal{B}$ with corresponding scores $\mathcal{S}$, greedy-NMS firstly selects the one $\mathcal{M}$ with the maximum score and moves it from set $\mathcal{B}$ to the set of final detections $\mathcal{F}$. It then removes any box in $\mathcal{B}$ and its score in $\mathcal{S}$ that has an overlap with $\mathcal{M}$ higher than a manually set threshold $N_t$. This process is repeated for the remaining $\mathcal{B}$ set.

Applying greedy NMS with a low threshold $N_t$ like 0.5 may increase the miss-rate, especially in crowd scenes. The reason lies in there may be many pairs of crowded objects which have higher overlaps than this suppressing threshold $N_t$. Within these pairs, when the proposal with the maximum score $\mathcal{M}$ is selected, all the surrounding detection boxes that have overlaps greater than $N_t$ are suppressed, including the nearby detections that actually locate the other ground truth instances. In this case, true positives may be removed after the NMS processing with a low $N_t$, increasing the miss rate.

Also, a high $N_t$ like 0.7 may increase false positives as many neighboring proposals that are overlapped often have correlated scores. Although more highly overlapped true positives can be kept, the increase in false positives may be more serious because the number of objects is typically smaller than the number of proposals generated by a detector. Therefore, using a high NMS threshold is not a good choice either.

To address this issue, the soft version of the greedy-NMS algorithm, \emph{i.e.} soft-NMS \cite{soft_nms},  writes the suppressing step as a re-scoring function:
\[
    s_i=
\begin{cases}
    s_i,& \textrm{iou}(\mathcal{M}, b_i) < N_t\\
    s_if(\textrm{iou}(\mathcal{M},b_i)),              &  \textrm{iou}(\mathcal{M}, b_i) \ge N_t
\end{cases}
\;,
\]
where $f(\textrm{iou}(\mathcal{M},b_i))$ is an overlap based weighting function to change the classification score $s_i$ of a box $b_i$ which has a high overlap with $\mathcal{M}$. According to this formulation, in greedy-NMS, $f(\textrm{iou}(\mathcal{M},b_i))\equiv0$, which means that $b_i$ should be directly removed. In soft-NMS, either $f(\textrm{iou}(\mathcal{M},b_i))=(1 - \textrm{iou}(\mathcal{M}, b_i))$ or $f(\textrm{iou}(\mathcal{M},b_i))=e^{-\frac{\textrm{iou}(\mathcal{M}, b_i)^2}{\sigma}}$ decays the scores of detections as an increasing function of overlap with $\mathcal{M}$. With the soft penalty, if $b_i$ contains another object not covered by $\mathcal{M}$, it does not lead to a miss at a lower detection threshold. However, as an increasing function, it still assigns a greater penalty to the highly overlapped boxes, which approximately equals to that in greedy-NMS.

Actually, both the design of greedy-NMS and soft-NMS follows the same hypothesis: the detection boxes with higher overlaps with $\mathcal{M}$ should have a higher likelihood of being false positives. This hypothesis has no problem when it is used in general object detection, as occlusions in a crowd rarely happen. However, this assumption does not hold in the crowded scenario, where human instances are highly overlapped with each other and should not be treated as false positives. Therefore, to adapt to pedestrian detectors in crowd scenes, NMS should take the following conditions into account,

\begin{itemize}
    \item For detection boxes which are far away from $\mathcal{M}$, they have a smaller likelihood of being false positives and they should thus be retained.

    \item For highly overlapped neighboring detections, the suppression strategy depends on not only the overlaps with $\mathcal{M}$ but also whether $\mathcal{M}$ locates in the crowded region. If $\mathcal{M}$ locates at the crowded region, its highly overlapped neighboring proposals are very likely to be true positives and should be assigned a lighter penalty or preserved. But for the instance in the sparse region, the penalty should be higher to prune false positives.

\end{itemize}

\subsection{Adaptive-NMS}
According to the above analysis, increasing the NMS threshold to preserve neighboring detections with high overlaps when the object is in a crowded region seems to be a promising solution to NMS in crowd scenes. It is also clear that the highly-overlapped proposals in the sparse region should be removed, as they are more likely to be false positives.

To quantitatively design the pruning strategy, we first define the object density as follows,
$$d_i \coloneqq \max\limits_{{b_j \in \mathcal{G}, i \neq j}} \textrm{iou}(b_i, b_j),$$
where the density of the object $i$ is defined as the max bounding box IoU with other objects in the ground truth set $\mathcal{G}$. The density of objects indicates the level of crowd occlusion.

With this definition, we propose to update the pruning step with the following strategy,
$$N_{\mathcal{M}}:=\max(N_t, d_{\mathcal{M}}),$$
\[
    s_i=
\begin{cases}
    s_i,& \textrm{iou}(\mathcal{M}, b_i) < N_{\mathcal{M}}\\
    s_if(\textrm{iou}(\mathcal{M},b_i)), &  \textrm{iou}(\mathcal{M}, b_i) \ge N_{\mathcal{M}}
\end{cases}
\;,
\]
where $N_{\mathcal{M}}$ denotes the adaptive NMS threshold for $\mathcal{M}$, and $d_{\mathcal{M}}$ is the density of the object $\mathcal{M}$ covers. We note three properties of this suppression strategy. (1) When the neighboring boxes which are far away from $\mathcal{M}$ (\emph{i.e.}, $\textrm{iou}(\mathcal{M}, b_i) < N_t$), they are retained the same as the original NMS does. (2) If $\mathcal{M}$ locates in the crowded region (\emph{i.e.}, $d_{\mathcal{M}}>N_t$), the density of $\mathcal{M}$ is used as the adaptive NMS threshold. Hence, the neighboring proposals are preserved, as they probably locate other objects around $\mathcal{M}$. (3) For the objects in a sparse region (\emph{i.e.}, $d_{\mathcal{M}} \le N_t$), the NMS threshold $N_{\mathcal{M}}$ equals to $N_t$. Then, the pruning step is equivalent to the original NMS, where very close boxes are suppressed as false positives.

The adaptive-NMS algorithm is formally described in Fig.~\ref{fig:algorithm}. As we only replace the fixed threshold $N_t$ with the adaptive ones, the computational complexity for adaptive-NMS is the same as traditional greedy-NMS and soft-NMS. The only extra cost for adaptive-NMS is an $N$-element list that stores the predicted density for each proposal, which is negligible for today's hardware configuration. Hence the adaptive-NMS does not affect the running time of current detectors much, keeping the efficiency as that of greedy-NMS and soft-NMS.

Note that adaptive-NMS works well with both greedy-NMS and soft-NMS. For fair comparison with soft-NMS, we adopt the original re-scoring function in greedy-NMS by default if not specified.

Once we know the density of the object, the adaptive-NMS flexibly preserves its neighbors and prunes the false positives. But we actually skip a major issue that is how to predict the density of each object, which is described in the next section.

\subsection{Density Prediction}
We treat density prediction as a regression task, where the target density value is calculated following its definition and the training loss is the Smooth-L1 loss.

A natural way for this regression is to add a parallel head layer at the top of the network just like classification and localization. However, the features used for detection only contain the information of the object itself, \emph{e.g.}, appearance, semantic feature and position. For density prediction, it is very difficult to estimate the density using the individual object information since it needs more clues about the surrounding objects.

To counter this, we design an extra subnet of three convolutional layers, as shown in Fig.~\ref{fig:detector}, to predict the density of each proposal. We note that this subnet is compatible with both the two-stage and one-stage detectors. For two-stage detectors, we construct the density subnet behind RPN. We first apply a $1\times 1$ conv layer to reduce the dimension of the convolutional feature maps, and we then concatenate the reduced feature maps as well as the objectness and bounding boxes predicted by RPN as the input of the density subnet. Moreover, we apply a large kernel ($5\times 5$) at the final conv layer of the density subnet to take the surrounding information into account. 
For one-stage detectors, the density subnet is deployed behind the final detection network in a similar way.

\section{Experiments}
To validate the proposed adaptive-NMS method, we conduct several experiments on two crowd pedestrian datasets: CityPersons \cite{citypersons} and CrowdHuman \cite{crowdhuman}.

\subsection{CityPersons}
\label{sec:citypersons}

\textbf{Dataset and Evaluation Metrics.} The CityPersons \cite{citypersons} dataset is a new pedestrian detection dataset which is built on top of the semantic segmentation dataset CityScapes \cite{cityscapes}. It records street views across 18 different cities in Germany with various weather conditions. The dataset includes 5, 000 images (2, 975 for training, 500 for validation and 1, 525 for testing) with $\sim$ 35, 000 labeled persons plus $\sim$ 13, 000 ignored region annotations. Both bounding box annotations of full bodies and visible parts are provided. Moreover, there are approximately 7 pedestrians in average per image, with 0.32 pairwise crowd instances (density higher than 0.5).

Following the evaluation protocol in CityPersons, all of our models on this dataset are trained on the reasonable training set and evaluated on the reasonable validation set. The log MR averaged over FPPI range of $[10^{-2}, 10^0]$ (MR$^{-2}$) is used to evaluate the detection performance (lower is better).

\textbf{Detector.} To demonstrate the effectiveness of adaptive-NMS, we conduct two types of baseline detectors.

For two-stage detectors, we generally follow the adapted Faster R-CNN framework \cite{citypersons} and use the pre-trained VGG-16~\cite{vgg} as the backbone. We also keep the same anchor sizes and ratios as in \cite{citypersons}. To improve the detection performance of small pedestrians, we adopt a common trick to use dilated convolution and the final feature map is $1/8$ of the input size.

For one-stage detectors, we modify RFB Net~\cite{RFB} and also use the VGG-16~\cite{vgg} pre-trained on ILSVRC CLSLOC~\cite{imagenet} as the backbone network. Besides, we follow the extension strategy in \cite{RFB} to up-sample the conv7\_fc feature maps and concat it with the conv4\_3 to improve the detection accuracy of pedestrians of small scales.

For fair comparison, we train the two base detectors with the density sub-network together. All the parameters in the new convolutional layers are randomly initialized with the MSRA method~\cite{MSRA}. We optimize both two detectors using Stochastic Gradient Descent (SGD) with 0.9 momentum and 0.0005 weight decay. For adapted Faster-RCNN, we train it on 4 Titan X GPUs with the mini-batch of 1 image per GPU. The learning rate starts at $10^{-3}$ for the first $20k$ iterations, and decays to $10^{-4}$ for another $10k$ iterations. For RFB Net, we set the batch size at 8 on 4 Titian X GPUs.  We also follow its ``warm-up'' strategy \cite{RFB} that gradually ramps up the learning rate from $10^{-6}$ to $2 \times 10^{-3}$, and then divide the learning rate by 10 at 120 and 180 epochs with totally 200 epochs in training.

\textbf{Ablation Study on Adaptive-NMS.} We first ignore the predicted densities and apply greedy-NMS and soft-NMS on detection results with various parameters. We search the NMS threshold $N_t$ in greedy-NMS and soft-NMS with the ``linear'' method to report the best results at $N_t=0.5$. We also try several normalizing parameters $\sigma$ in soft-NMS using the ``Gaussian'' method, but they all increase the miss rate by about 1\%. We thus only report the ``linear'' results for clear presentation in the rest of the paper. We also report the total recall and Average Precision (AP) on the \emph{Reasonable} set for more reference.

As shown in Table~\ref{table:ablation}, using the traditional greedy-NMS, the adapted Faster R-CNN detector achieves 14.5\% MR$^{-2}$ on the validation set, which is slightly better than the reported result (15.4\% MR$^{-2}$) in \cite{citypersons}. The RFB Net detector achieves 13.9\% MR$^{-2}$, which is slightly better than the current single-shot detectors \cite{haikang} in CityPersons.

The soft-NMS with the ``linear'' method slightly reduces the MR$^{-2}$ by 0.3\% (\emph{i.e.}, 14.2\% MR$^{-2}$ \emph{vs.} 14.5\% MR$^{-2}$) for Faster R-CNN detector. For RFB Net, soft-NMS does not work well. Combining adaptive-NMS with soft-NMS also has minor or even negative improvements on metric MR$^{-2}$. The reason is that the low-score detections soft-NMS keeps could be out of the right-hand boundary of FPPI range $[10^{-2}, 10^0]$. So MR$^{-2}$ does not benefit from it.

With the proposed adaptive-NMS method, the MR$^{-2}$ score of the Faster R-CNN detector significantly drops to 12.9\% with a 1.6\% reduction, and that of the RFB Net detector also reduces by 1.2\% (\emph{i.e.}, 13.9\% MR$^{-2}$ \emph{vs.} 12.7\% MR$^{-2}$). These results indicate that adaptive-NMS keeps more true positives, and it is a more effective post-processing algorithm for detecting pedestrians in crowded scenarios.

\begin{table}[t]
\begin{center}
\scalebox{.47}{
\resizebox{\textwidth}{!}{
\begin{threeparttable}
\begin{tabular}{l|ccc|c|c|c|c}
\multicolumn{4}{c|}{\multirow{2}{*}{Method}}                                                                 & \multirow{2}{*}{Backbone} & \multicolumn{3}{c}{\emph{Reasonable}}  \\ \cline{6-8}
\multicolumn{4}{c|}{}                                                                                        &                           & \multicolumn{1}{l|}{MR$^{-2}$} & Recall & AP \\ \hline
\multicolumn{4}{c|}{Faster RCNN \cite{citypersons} (two-stage)}
& VGG-16                    & 15.4                    & -    & -  \\
\multicolumn{4}{c|}{TLL \cite{haikang} (one-stage)}                                                                 & ResNet-50                 & 14.4                    & -    & -  \\ \hline
                                                                         & greedy    & soft      & adaptive  &                           &                         &        \\ \hline
\multirow{3}{*}{\begin{tabular}[c]{@{}l@{}}Faster \\ R-CNN\end{tabular}} & \checkmark &           &           & VGG-16                    & 14.5                    & 95.6   & 93.8  \\
                                                                         &           & \checkmark &           & VGG-16                    & 14.2                    & 98.3   &  94.9\\
                                                                         &           &           & \checkmark & VGG-16                    & \textbf{12.9}                    & 97.7     &  \textbf{95.3}\\
                                                                         &           & \checkmark  & \checkmark & VGG-16                    & 14.1& \textbf{98.4}     &  95.0 \\
                                                                         \hline
\multirow{3}{*}{RFB Net}                                                 & \checkmark &           &           & VGG-16                    & 13.9                    & 95.6   & 94.3  \\
                                                                         &           & \checkmark &           & VGG-16                    & 14.2                    & \textbf{99.2}  &94.1  \\
                                                                         &           &           & \checkmark & VGG-16                    & \textbf{12.7}                    & 97.4  & \textbf{95.0}   \\
                                                                         &           & \checkmark& \checkmark & VGG-16                    & 14.3                    & \textbf{99.2}  &  94.1

\end{tabular}
\end{threeparttable}}}
\end{center}
\caption{Ablation study for greedy-NMS, soft-NMS and adaptive-NMS. We only report the best results of greedy-NMS and soft-NMS with 0.5 NMS threshold for clear comparison.}
\label{table:ablation}
\end{table}

\begin{figure}[t]
\begin{center}
   \includegraphics[width=0.95\linewidth]{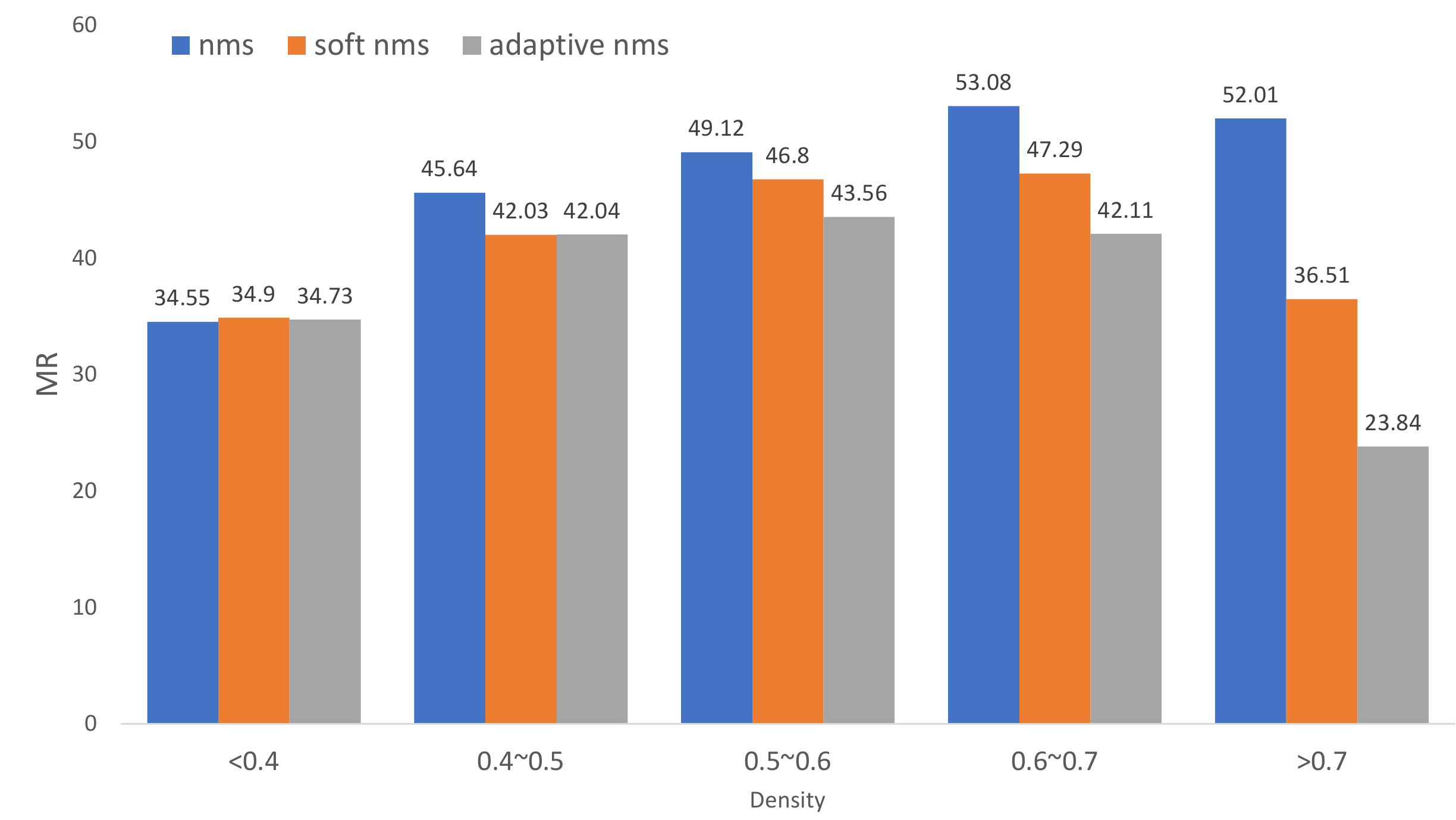}
\end{center}
\vspace{-0.3cm}
   \caption{The MR$^{-2}$ results in 5 groups with different levels of crowd occlusions. Adaptive-NMS works much better on the higher density groups.}
\label{fig:analysis}
\vspace{-0.4cm}
\end{figure}

\begin{figure}[thbp]
\begin{center}
   \includegraphics[width=0.90\linewidth]{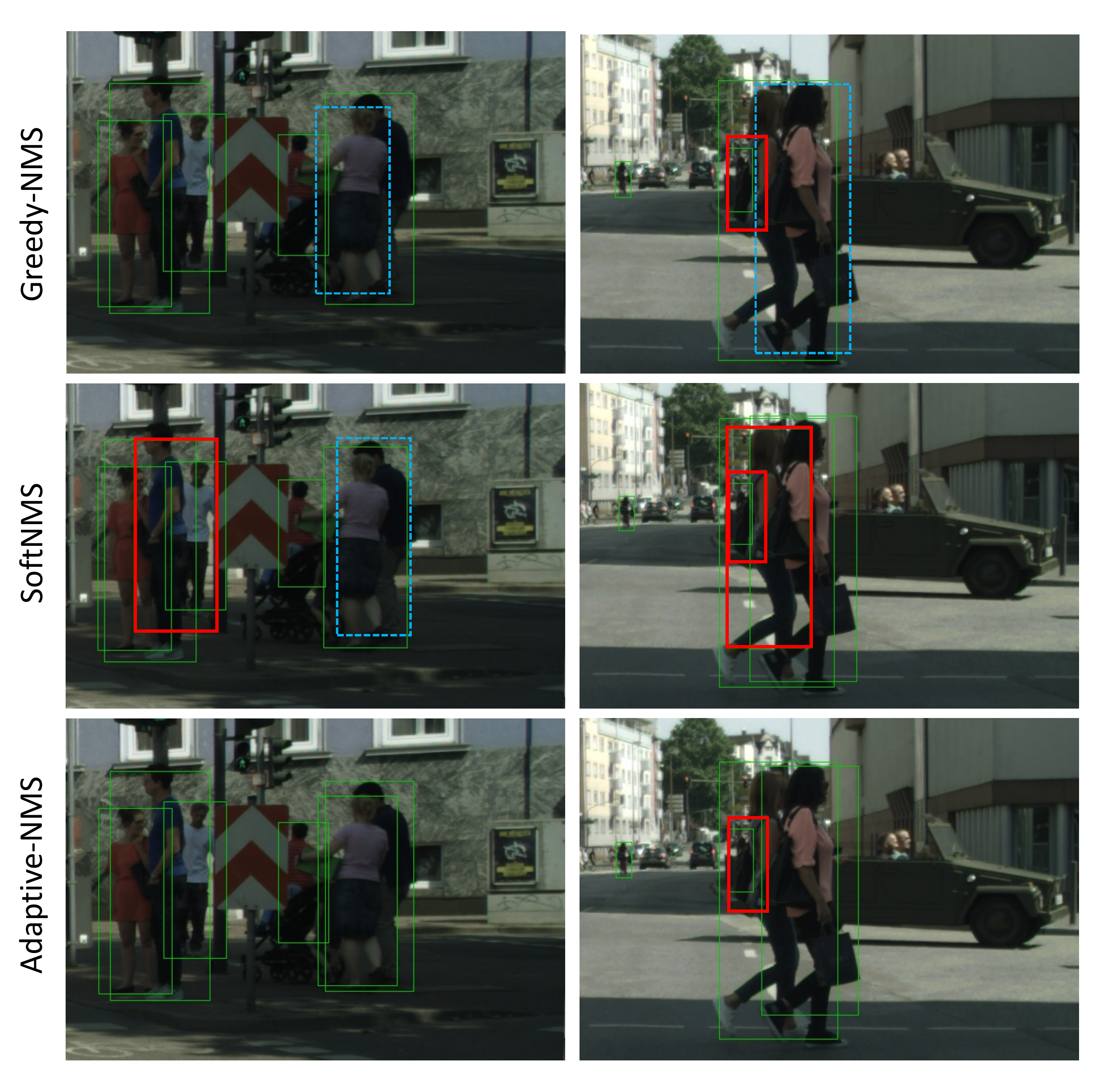}
\end{center}
\vspace{-0.4cm}
   \caption{Visual comparisons of the Faster R-CNN pedestrian prediction results (green boxes) with greedy-NMS, soft-NMS and adaptive-NMS. Blue boxes are missing objects, while red boxes are false positives. The scores thresholded for visualization are above 0.3.}
\label{fig:cases}
\vspace{-0.5cm}
\end{figure}

\begin{table*}[thbp]
\begin{center}
\scalebox{.73}{
\resizebox{\textwidth}{!}{
\begin{threeparttable}
\begin{tabular}{ccc|c|c|c|ccc}
\multicolumn{3}{c|}{Method}                                  & Scale       & Backbone  & \emph{Reasonable} & \emph{Heavy} & \emph{Partial} & \emph{Bare} \\ \hline
\multicolumn{3}{c|}{\multirow{2}{*}{Adapted Faster RCNN \cite{citypersons}}}   & $\times1$   & VGG-16    & 15.4       & -     & -       & -    \\
\multicolumn{3}{c|}{}                                       & $\times1.3$ & VGG-16    & 12.8       & -     & -       & -    \\
\multicolumn{3}{c|}{\multirow{2}{*}{Repulsion Loss \cite{repulsion}}}        & $\times1$   & ResNet-50 & 13.2       & 56.9  & 16.8    & 7.6  \\
\multicolumn{3}{c|}{}            & $\times1.3$ & ResNet-50 & 11.6       & 55.3  & 14.8    & 7.0  \\
\multicolumn{3}{c|}{\multirow{2}{*}{OR-CNN \cite{or_cnn}}}                & $\times1$   & VGG-16    & 12.8       & 55.7  & 15.3    & 6.7  \\
\multicolumn{3}{c|}{}                                       & $\times1.3$ & VGG-16    & 11.0       & 51.3  & 13.7    & 5.9  \\ \hline
\hline
\multicolumn{1}{c|}{}            & AggLoss \cite{or_cnn}  & Adaptive-NMS &             &           &            &       &         &      \\ \cline{1-3}
\multicolumn{1}{c|}{Faster RCNN}            &           & \checkmark    & $\times1$   & VGG-16    & 12.9      & 56.4  & 14.4    & 7.0  \\
\multicolumn{1}{c|}{} & \checkmark &              & $\times1$   & VGG-16    & 13.2       & 56.0  & 14.0    & 7.7  \\
\multicolumn{1}{c|}{}            & \checkmark & \checkmark    & $\times1$   & VGG-16    & \textbf{11.9}       & 55.2  & 12.6    & 6.2  \\
\multicolumn{1}{c|}{}            & \checkmark &              & $\times1.3$ & VGG-16    &  11.4       & 55.6  &         11.9    & 6.2  \\
\multicolumn{1}{c|}{}            & \checkmark & \checkmark    & $\times1.3$ & VGG-16    & \textbf{10.8}       & 54.0  &         11.4    & 6.2  \\ \hline
\multicolumn{1}{c|}{RFB Net}     &           & \checkmark    & $\times1$   & VGG-16    & 12.7        &  51.9  &         11.7    & 7.6  \\
\multicolumn{1}{c|}{}            & \checkmark &              & $\times1$   & VGG-16    & 13.1        &  51.7 &         12.0 & 7.4      \\
\multicolumn{1}{c|}{}            & \checkmark & \checkmark    & $\times1$   & VGG-16    & \textbf{12.0}        & 51.2  &    11.9   &  6.8
\end{tabular}
\end{threeparttable}}}
\end{center}
\vspace{-0.2cm}
\caption{Comparison of detection performance on the CityPersons validation set.}
\label{table:compare}
\vspace{-0.2cm}
\end{table*}

\textbf{Analysis.} The average log MR and recall on the \emph{reasonable} validation set do not explain us clearly where adaptive-NMS obtains significant gains in performance. We further divide the pedestrians with at least 50 pixel height in the validation set into 5 subsets according to their density (density $\le$ 0.4, 0.4 $<$ density $\le$ 0.5, 0.5 $<$ density $\le$ 0.6, 0.6 $<$ density $\le$ 0.7, density $>$ 0.7). For better demonstration, we compare the results of Faster R-CNN with greedy-NMS, soft-NMS (``linear'') as well as adaptive-NMS on these subsets. From Fig.~\ref{fig:analysis}, we can infer that for sparse pedestrians whose density is less than 0.4, all the three NMS algorithms show similar performance. When the density increases, the proposed adaptive-NMS significantly reduces the miss rate compared with the two counterparts. This demonstrates that adaptive-NMS performs better-post processing in the crowd scene, keeping more highly-overlapped true positives.

In addition, we also show some visual results of the Faster R-CNN detector with greedy-NMS, soft-NMS and adaptive-NMS  for comparison. As Fig.~\ref{fig:cases} shows, adaptive-NMS keeps more crowded true positives and still removes false positives in the sparse region at the same time.

\begin{figure}[t]
\begin{center}
   \includegraphics[width=0.8\linewidth]{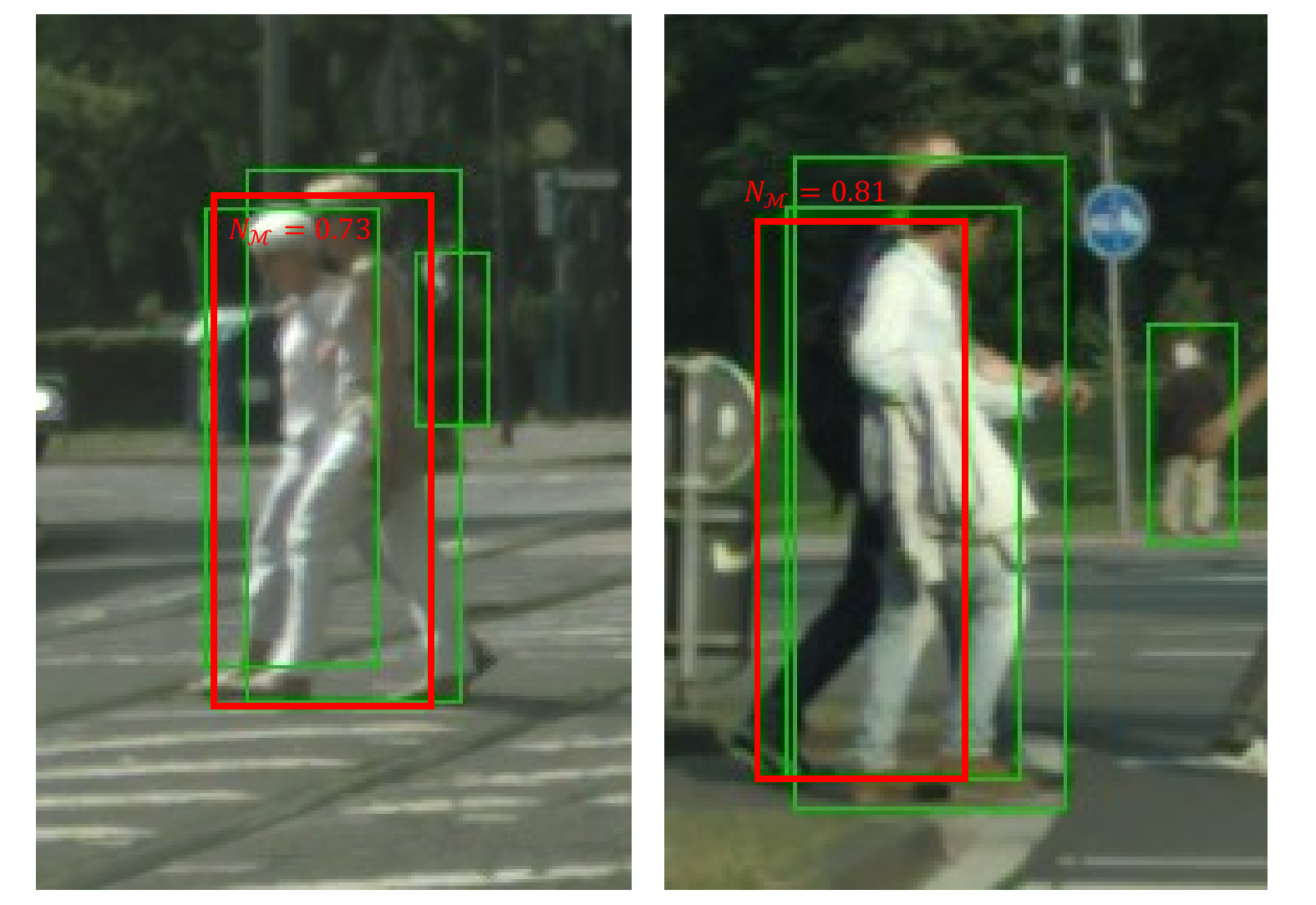}
\end{center}
\vspace{-0.3cm}
   \caption{Failure cases of adaptive-NMS with the 0.3 visual score threshold. Red boxes are false positives. As the NMS threshold ($N_{\mathcal{M}}$) increases for crowd instances, more false positives are also preserved if the proposals are not compact.}
\label{fig:failure}
\vspace{-0.1cm}
\end{figure}

\textbf{Comparison to the State-of-the-art.} As adaptive-NMS only focuses on the post process of detectors, it conveniently works with typical advanced pedestrian detectors. Moreover, as illustrated in Fig.~\ref{fig:failure}, the minor punishment in the crowd instances increases false positives if the proposals of the ground-truth objects are not compact. \emph{}Hence, to better validate the effectiveness of adaptive-NMS, we follow \cite{or_cnn} to add the AggLoss term on the regression loss to enforce the proposals locate closely and compactly to the ground-truth, which is defined as
\[
\begin{array}{ll}
{\cal L}_{\text{com}}(\{t_i\}, \{t_i^\ast\})=\frac{1}{N_{\text{com}}}\sum_{i=1}^{N_{\text{com}}}\Delta(\tilde{t}_i^\ast - \frac{1}{|\Phi_i|}\sum_{j\in{\Phi_i}}t_j),
\end{array}
\]
where $N_{\text{com}}$ is the total number of ground truths associated with more than one anchor, $|\Phi_i|$ is the number of anchors associated with the $i$-th ground truth object, $\tilde{t}_i^\ast$ and $t_i$ are the associated coordinates of the ground truth and proposals.

In Table~\ref{table:compare}, we follow the strategy in \cite{repulsion} and \cite{or_cnn} to divide the \emph{Reasonable} subset (occlusion $<$ 35\%) in the validation set into the \emph{Partial} (10\% $<$ occlusion $<$ 35 \%) and  \emph{Bare} (occlusion $\le$ 10\%) subsets. Meanwhile, we denote the pedestrians with the occlusion ratio of more than 35\% as the \emph{Heavy} set. With the $\times1$ scale of input images, adaptive-NMS improves the baseline detectors to reach comparable results with those of other counterpart pedestrian detectors without any additional module. For Faster R-CNN, when we add AggLoss~\cite{or_cnn} with adaptive-NMS, it achieves the state-of-the-art results on the validation set of CityPersons by reducing 0.9\% MR$^{-2}$ (\emph{i.e.}, 11.9\% \emph{vs.} 12.8\% of \cite{or_cnn}). For RFB Net, adaptive-NMS with AggLoss also pushes the performance to 12.0\% MR$^{-2}$.

We then enlarge the size of the input image as in \cite{repulsion,citypersons,or_cnn}. Due to the GPU memory issue, we do not train the RFB Net detector with $\times1.3$ scale of input size. For Faster R-CNN,  it achieves the best performance of 10.8\% MR$^{-2}$. In addition, we also evaluate the proposed Adaptive-NMS method on the testing set of CityPersons and report the results in Table~\ref{table:test}. With $\times1.3$ scale and AggLoss, the Faster R-CNN detector achieves 11.79\% MR$^{-2}$, while Adaptive-NMS further improves the result to 11.40\% MR$^{-2}$. It is worth noting that other counterparts either employ a part occlusion-aware pooling module \cite{or_cnn} or a stronger backbone network \cite{repulsion} (\emph{i.e,}, ResNet-50). As adaptive-NMS has few constraints for the architecture of detectors, we believe the performance of adaptive-NMS can be further improved with these techniques.

\begin{table}[thbp]
\begin{center}
\scalebox{.47}{
\resizebox{\textwidth}{!}{
\begin{threeparttable}
\begin{tabular}{l|c|c|c}
Method                          & Backbone  & Scale & Reasonable \\ \Xhline{1.2pt}
Adapted FasterRCNN~\cite{citypersons}              & VGG-16    & $\times$1.3   & 12.97      \\
Repulsion Loss~\cite{repulsion}                  & ResNet-50 & $\times$1.5   & 11.48      \\
OR-CNN~\cite{or_cnn}                          & VGG-16    & $\times$1.3   & 11.32      \\ \Xhline{1.2pt}
FasterRCNN+AggLoss              & VGG-16    & $\times$1.3   & 11.79      \\
FasterRCNN+AggLoss+Adaptive-NMS & VGG-16    & $\times$1.3   & 11.40
\end{tabular}
\end{threeparttable}}}
\end{center}
\caption{Comparison of detection performance on CityPersons test.}
\label{table:test}
\end{table}

\subsection{CrowdHuman}
\label{sec:crowdhuman}
\begin{table}[thbp]
\begin{center}
\scalebox{.45}{
\resizebox{\textwidth}{!}{
\begin{threeparttable}
\begin{tabular}{c|ccc}
                     & Caltech~\cite{caltech} & City~\cite{citypersons} & Crowd~\cite{crowdhuman} \\ \hline
\# person/img        & 0.32    & 6.47        & \textbf{22.64}      \\ \hline
\# pair/img          &         &             &            \\
iou\textgreater{}0.3 & 0.06    & 0.96        & 9.02       \\
iou\textgreater{}0.5 & 0.02    & 0.32        & \textbf{2.40}       \\
iou\textgreater{}0.7 & 0.00    & 0.08        & 0.33
\end{tabular}
\end{threeparttable}}}
\end{center}
\vspace{-0.2cm}
\caption{Comparison in terms of the average number of persons and pair-wise overlap between two instances on the three datasets.}
\label{table:crowdhuman}
\vspace{-0.2cm}
\end{table}

\textbf{Dataset and Evaluation Metrics.} Recently, CrowdHuman \cite{crowdhuman} has been released to specifically target to the crowd issue in the human detection task. It collects 15, 000, 4, 370 and 5, 000 images from the Internet for training, validation and testing respectively. There are $\sim340k$ persons and $\sim99k$ ignore region annotations in the training set. Moreover, the CrowdHuman dataset is of much higher crowdedness compared with all the previous ones (\emph{e.g.}, CityPersons \cite{citypersons}, KITTI~\cite{kitti} and Caltech~\cite{caltech}). As shown in Table~\ref{table:crowdhuman}, it contains approximately 22.6 pedestrians in average per image as well as 2.4 pairwise crowd instances (density higher than 0.5).

We follow the evaluation metric used in CrowdHuman \cite{crowdhuman}, denoted as MR$^{-2}$ as introduced in Section~\ref{sec:citypersons}. All the experiments are trained in the CrowdHuman training set and evaluated in the validation set, and only the full body region annotations are used for training and evaluation.

\textbf{Detector.} We also conduct two baseline detectors to evaluate the performance of adaptive-NMS.

For two-stage detectors, as Faster-RCNN \cite{citypersons} with the VGG-16 backbone fails to reach a good baseline result in our early experiments, we follow \cite{crowdhuman} to employ the Feature Pyramid Network (FPN) \cite{FPN} with a ResNet-50 \cite{ResNet} as the new backbone network. We also use the same settings of design parameters, such as [1.0,1.5,2.0,2.5,3.0] anchor ratios and no clipping proposals. For one-stage detectors, we use RFB Net with the same architecture as in Section~\ref{sec:citypersons}.

As the images of CrowdHuman are collected from websites with various sizes, we resize them so that the shorter image side is 800 pixels for FPN. The input size of RFB Net is set as 800 $\times$ 1200. The base learning rate is set to 0.02 and 0.002 for FPN and RFB Net respectively, and divided by 10 at $150k$ and $450k$ for FPN, and $400k$ and $600k$ for RFB Net. The SGD solver with 0.9 momentum is adopted to optimize the networks on 4 Titian X GPUs with the mini-batch of 2 images per GPU, while the weight decay is set at 0.0001 and 0.0005 for FPN and RFB Net respectively. For fair comparison with \cite{crowdhuman}, we do not use additional losses such as AggLoss \cite{or_cnn} or Repulsion Loss \cite{repulsion}.

\textbf{Evaluation Results.} In Table~\ref{table:crowdhuman-results}, our baseline detectors achieve comparable results as \cite{crowdhuman} does. When we replace greedy-NMS with adaptive-NMS, the miss rate drops by 2.62\% MR$^{-2}$ and 2.19\% MR$^{-2}$ for FPN and RFB Net respectively. It proves that the proposed adaptive-NMS algorithm is effective and has a good potential for processing detectors in crowd scenes.

\begin{table}[thbp]
\begin{center}
\scalebox{.45}{
\resizebox{\textwidth}{!}{
\begin{threeparttable}
\begin{tabular}{l|ccc|c|c|c}
          & greedy & soft & adaptive               & MR$^{-2}$  &Recall &AP  \\ \hline
FPN~\cite{crowdhuman} & \checkmark  &        &     & 50.42 & 90.24 & 84.95\\ \hline
FPN       & \checkmark  &           &              & 52.35 & 90.57 & 83.07\\
          &            & \checkmark &              & 51.97 & 91.73 & 83.92\\
          &            &           & \checkmark    & \textbf{49.73} & 91.27 & 84.71\\ \hline
RetinaNet~\cite{crowdhuman} & \checkmark  &           &              & 63.33 & 93.80 & 80.83\\ \hline
RFB Net   & \checkmark  &           &              & 65.22 & 94.13 & 78.33\\
          &            & \checkmark &              & 66.34 &95.37 & 78.10\\
          &            &           & \checkmark    & \textbf{63.03} & 94.77 & 79.67
\end{tabular}
\end{threeparttable}}}
\end{center}
\vspace{-0.2cm}
\caption{Evaluation of full body detections on the CrowdHuman validation set.}
\label{table:crowdhuman-results}
\vspace{-0.4cm}
\end{table}
\section{Conclusions}
In this paper, we present a new adaptive-NMS method to better refine the bounding boxes in crowded scenarios. Adaptive-NMS applies a dynamic suppression strategy, where an additionally learned sub-network is designed to predict the threshold according to the density for each instance.  Experiments are conducted on the CityPersons \cite{citypersons} and CrowdHuman \cite{crowdhuman} databases, and state of the art results are reached, showing its effectiveness.

\section*{Acknowledgment}
This work is funded by the National Key Research and Development Plan of China under Grant 2016YFC0801002 and the Research Program of State Key Laboratory of Software Development Environment.

{\small
\bibliographystyle{ieee}
\bibliography{cvpr}
}

\end{document}